# Context Aware Image Annotation in Active Learning

Yingcheng Sun and Kenneth Loparo

Case Western Reserve University, Cleveland, OH 44106, USA
{yxs489,kal4}@case.edu

**Abstract.** Image annotation for active learning is labor-intensive. Various automatic and semi-automatic labeling methods are proposed to save the labeling cost, but a reduction in the number of labeled instances does not guarantee a reduction in cost because the queries that are most valuable to the learner may be the most difficult or ambiguous cases, and therefore the most expensive for an oracle to label accurately. In this paper, we try to solve this problem by using image metadata to offer the oracle more clues about the image during annotation process. We propose a Context Aware Image Annotation Framework (CAIAF) that uses image metadata as similarity metric to cluster images into groups for annotation. We also present useful metadata information as context for each image on the annotation interface. Experiments show that it reduces that annotation cost with CAIAF compared to the conventional framework, while maintaining a high classification performance.

## 1    Introduction

Digital photos are now part of our everyday life due to the popularization of digital cameras, smartphones, surveillance systems, and other image capture devices [1, 2]. The number of photos and pictures taken per day increases every year. In semantic image classification or Content-Based Image Retrieval (CBIR) tasks such as face recognition and automatic pilot, a large amount of labeled data is necessary in the form of a training set, and it will entail significant manual effort. Hence, developing a strategy to minimize human annotation effort in a multi-label problem is of paramount practical importance. Though various automatic or semi-automatic annotation techniques are proposed [3, 4], the results are still not satisfactory and convincing enough [5], so manual annotation is inevitable at the present stage.

   Active learning algorithms iteratively query only the most informative instances to label have gained popularity to reduce human annotation effort. When exposed to large quantities of unlabeled data, such algorithms automatically select the promising and exemplar instances to be labeled manually. This tremendously reduces the annotation effort and also endows the model with greater generalization capability as it gets trained on the salient examples from the underlying data population [6]. In most applications, batch mode active learning, where a set of items is picked all at once to be labeled and then used to re-train the classifier, is most feasible because it does not require the model to be re-trained after each individual selection and makes most efficient use of human labor for annotation [7].

Most previous work focuses on developing the strategies of selecting samples, but the way of querying labels from the annotators are seldom discussed. Burr et al. [8] proposed that minimizing the number of queries does not guarantee the reduction of the whole annotation cost because the queries that are most valuable to the learner may be the most difficult or ambiguous cases, and therefore the most expensive for an oracle to label accurately. Figure 1 shows such an example of image annotation of the Statue of Liberty.

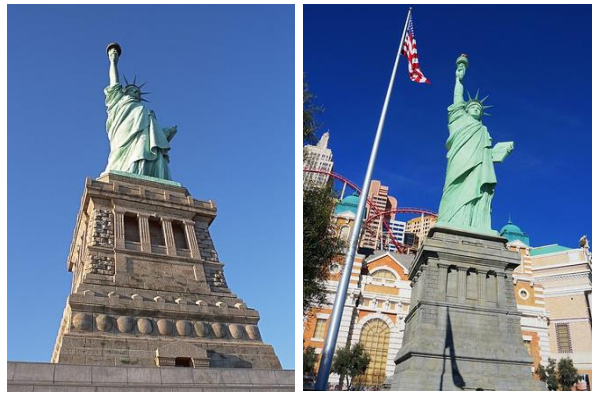

**Fig. 1.** Left: the original Statue of Liberty in New York City. Right: the replica in Las Vegas.

Given such two images in monument recognition [9] or landmark classification [10] task, it is very hard to annotate them correctly if the oracle does not know there is a replica of Statue of Liberty in Las Vegas, since the photos in Figure 1 depict almost the same visual objects. Even with some background knowledge, it still takes time for the oracle to tell them apart because of the uncertainty. However, the image metadata such as the geographical location can help the oracle to annotate them fast and accurate enough in this case.

In traditional active learning frameworks with batch mode, a group of images are picked up and shown to annotators without any specific order. Sometimes the class label of images switch frequently during annotation process, which might increase the annotation time and error rate. Like the example shown in Figure 2, the left image might be wrong annotated on its own without any context because it might be a flower petal, but it could also be a piece of fruit or possibly an octopus tentacle which is very ambiguous. However, in the context of a neighborhood of images (the right column) with similar metadata like taken time, author or user tags, it is clearer that the left one shows a flower. The context of additional unannotated images disambiguates the visual classification task [11].

To address the above issues, we proposed a Context Aware Image Annotation Framework (CAIAF) [34]. In this paper, we will discuss more details about using image metadata as context information to organize images during annotation process.

Most images on the web carry metadata; the idea of using it to improve visual classification is not new. Prior work takes advantage of user tags for image classification and retrieval [12, 13], uses GPS data [14, 15] to improve image classification, and utilizes timestamps [16] to both improve recognition and study topical evolution over time. The motivation behind much of this work is the notion that images with similar metadata tend to depict similar scenes.

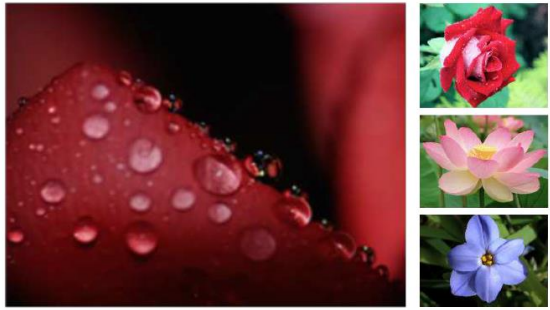

**Fig. 2.** Left: a flower petal. Right: neighbors of the left one in terms of metadata.

In CAIAF, similar images display together in each batch after clustering by the assigned metadata, and useful metadata information of the image to be labeled is also presented on the annotation interface. By doing these, annotators will have more clues about each image during the annotation process and thus reduce the annotation cost and improves the performance. Experiments show that CAIAF saves the annotation time and also leads to a better annotation performance.

The rest of this paper is organized as follows. In Section 2, we present related work. In Section 3, we propose the CAIAF and explain the metadata used in this framework. In Section 4, we introduce the dataset, comparison methods and evaluation metrics, as well as the experimental results. We conclude our work in Section 5.

## 2 Related Work

### 2.1 Improve Image Classification by Exploiting Metadata

Most images on the web carry metadata that can be very useful to improve image visual classification. One class of image metadata where this notion is particularly relevant is social-network metadata, which can be harvested for images embedded in social networks such as Flickr. In [12] the authors study the relationship between tags and manual annotations, with the goal of recovering annotations using a combination of tags and image content. The problem of recommending tags was studied in [17], where possible tags were obtained from similar images and similar users. The same problem was studied in [18], who exploit the relationships between tags to suggest future tags based on existing ones. Friendship information between users was studied for tag recommendation in [19], and in [20] for the case of Facebook. McAuley and Leskovec [21] pioneered the study of multilabel image annotation using metadata, and demonstrated impressive results using only metadata and no visual features. Justin et al. [11] exploits social-network metadata to improve image annotation.

Another commonly used source of metadata comes directly from the camera, in the form of Exif and GPS data [10, 22, 23]. Such metadata can be used to determine whether two photos were taken by the same person, or from the same location, which provides an informative signal and context for certain image categories. Kevin et al. [24] use the GPS coordinates as location context to improve image classification. Matthew et al. [25] integrates Exif metadata like exposure time, flash use and subject distance to tackle the "indoor-outdoor" image classification problem.

These researches discuss different methods to build a better image classification model. Our work differs from them because we focus on the annotator (oracle) side, and tries to reduce the annotation cost by using image metadata.

### 2.2 Reduce Image Annotation Cost in Active Learning

Machine learning methods such as active learning, distant learning and reinforcement learning are widely used in classification tasks [35- 39]. Most previous work in active learning has assumed a fixed cost for acquiring each label, i.e., all queries are equally expensive for the oracle. Burr et al. [8] prove that the cost is not fixed, and they make an empirical study of annotation costs in four real-world text and image domains. They predicted the annotation cost in the text domain but failed in the image domain. Burr et al. later [26] propose an annotation paradigm DUALIST that solicits and learns from labels on both features and instances. It is fast enough to support real-time interactive speeds in text field. Qiang et.al [27] explored the possible factors are associated with the cost of time in clinical text annotation. Stefan et al. [28] discussed the "difficulty" of tweet that affects labeling performance of annotators. However, the problem of how to reduce the image annotation cost has not been studied.

Some researchers try to reduce the human annotation effort in active learning by using the current learned model to assist in the labeling of query instances in structured-output tasks like parsing [29] or named entity recognition [30]. Haertel, et al. [31] proposed a parallel active learning method which can eliminate the wait time with minimal staleness. Thiago et al. [32] introduce a ranked batch-mode active learning framework to reduce the manual labeling delays. However, these methods do not actually represent or reason about costs. Instead, they attempt to reduce the number of annotation actions required for a query. Our research tries to solve this problem from another perspective that uses metadata to give annotators more context, and then to reduce the annotation cost and improve the performance.

### 3 Context Aware Image Annotation Framework

Traditional active learning frameworks query the oracle one instance per time to label, even in batch mode, because the selected images are usually shown in an order of their informative vale or just randomly. It will not help the annotation too much if the oracle knows the previous labeled and following unlabeled images. However, we design a Context Aware Image Annotation Framework (CAIAF) by using the image metadata, and try to give annotators more clues in the annotation activity. In CAIAF, each batch of images are clustered by the similarity of their metadata and displayed in groups. Fig. 3 shows the process.

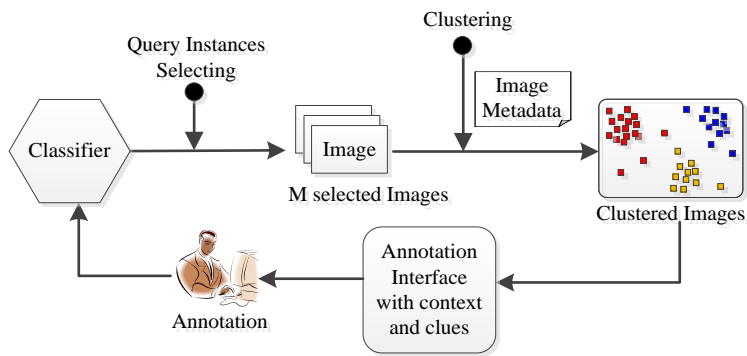

**Fig. 3.** The framework of labeling images with metadata as context information

First, *M* images are selected from corpus by query instances selecting algorithm with batch mode. Next, the selected images are clustered by the similarity of metadata. In this paper, we use K-means as the clustering method. Next, the clustered images will be shown to the annotator by groups. Each image is displayed with its metadata information. This process iterates until the threshold is met. In order to support a context description relevant for annotating photos, we have defined four context dimensions: location, time, user tags, and camera tags.

**Location**  With the widespread availability of cellphones and cameras that have GPS capabilities, it is common for images being uploaded to the Internet today to have GPS coordinates associated with them. With this geographical information in hand, it is much easier to correctly deduce the label of geo-related images, like the example of Statue of Liberty shown in Figure 1. Images will be clustered by their geodesic distance if the location is set as the context clue. The real location like "New York City" transferred from coordinates will also be shown on the annotation interface.

**Time**  The creation time (and date) of the photo is another dimension that can be used to organize the images. It allows the association of an instant (date and time) with a photo, and also of the different time interpretations and attributes listed above (e.g., night, Monday, July). Thus, the temporal concept can be used to cluster photographs by events such as sunrise and sunset. Images will be grouped by closeness of their timestamps if the time is set as the context clue. The time will also be shown on the annotation interface.

**User Tags**  One class of image metadata where this notion is particularly relevant is social-network metadata, which can be harvested for images embedded in social networks such as Flickr. These metadata, such as user-generated tags are applied to images by people as a means to communicate with other people; as such, they can be highly informative as to the semantic contents of images. We compute the distance between images using word embedding since it can capture the semantic similarity between texts [40, 41, 42]. Images will be clustered by their similarity if the "user tags" is set as the context. The original user- generated tags location will display the annotation interface.

**Camera Tags**   Exif metadata recorded by the camera provides cues independent of the scene content that can be exploited to improve image annotation. The Exif metadata standard for JPEG images includes a number of tags related to picture taking conditions, including FlashUsed, FocalLength, ExposureTime, Aperture, FNumber, ShutterSpeed, and Subject Distance. It is clear that some of these cues can help discriminate between certain scene types (e.g., long subject distances occur primarily on landscape photos); the scene classification problem at hand determines which cues help the most. We do not use the camera tag metadata as context in our paper because it is not a distinguishable feature for our dataset, but it has been proved that the scene brightness, subject distance and flash are salient in the problem of "indoor-outdoor" classification problem [25].

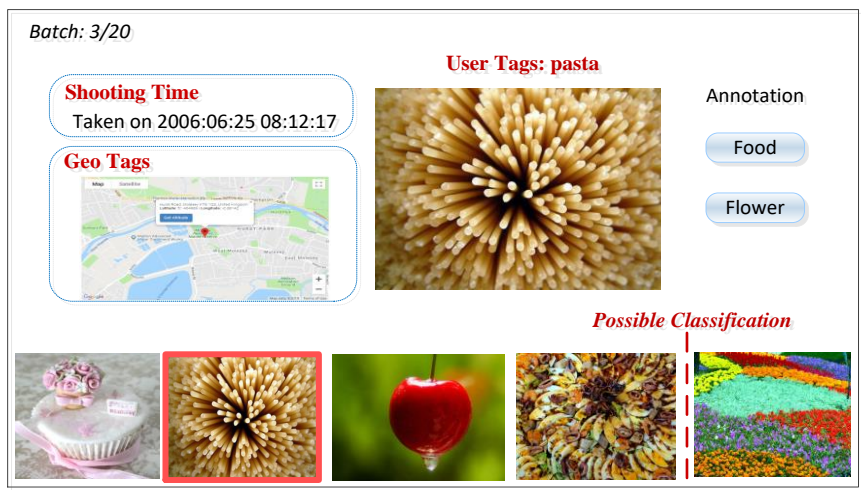

**Fig. 4.** A screenshot of the annotation interface developed based on CAIAF

A simple annotation interface is developed based on the proposed framework CAIAF. A screenshot of it is shown in Figure 4. In this interface, a batch of images are all presented to the oracle. In our experiment, we set the batch size as five, so there are five images listed in a row. The oracle needs to label them one by one, and choose the class of the current image with the button on the right corner. After clicking any of the class buttons, the next image will be chosen and shown on the upper part. In Figure 4, the "food or flower" image classification is being queried for annotation. The left column lists the temporal and geographical information of the current image being labeled. The "user tags" are listed the top of the image that can help to understand the content of the image. Besides these clues, the row of images is also clustered by the metadata, and partitioned by a dashed red line. With all these information as context, it is easier for the oracle to annotate the image fast and accurate. The number of batch of images that have been annotated and the total number of batches are shown on the left corner. The annotation time of each batch of images is logged.

## 4 Experiment And Analysis

### 4.1 Dataset

In this paper, we use the NUS-WIDE dataset [43] for our experiments. This dataset is created by NUS's Lab for Media Search and has been widely used for image labeling and retrieval. It consists of 269,648 images collected from Flickr with plentiful metadata, each manually annotated for the presence or absence of 81 labels. To make it easier for the annotation experiment, we picked 8 categories from them and make four pairs of comparison sets: bird and cat, flower and food, lake and ocean, town and temple and use five types of metadata information: image description, data and time, geographical coordinates, headline, and keywords as shown in Fig. 5. We discard images which metadata is not complete or unavailable. Following [44] we also discard images that labels are absent. We randomly picked 100 images from the left ones for each category, and 800 in total.

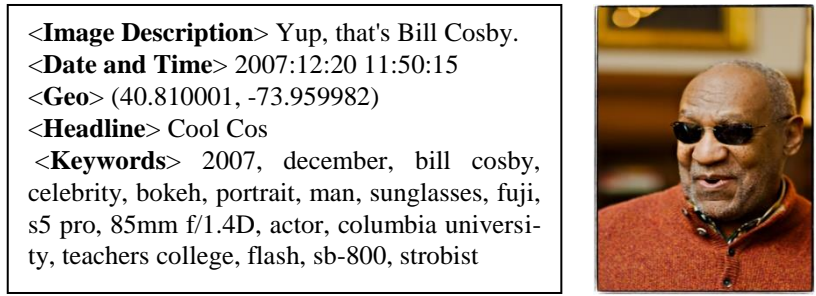

**Fig. 5.** Selected pieces of metadata information and the raw image

We developed a conventional image annotation interface without any metadata clues as the comparison. Image is queried one by one in that interface. We call it "plain interface" in this paper.

### 4.2 Experiment

We use a python active learning module[1] offered by Google for the image annotation experiment. We set the batch size as 5, and choose "Informative and diverse" as the active learning method and "Linear SVM" as the classification model. We choose the "image description" and "keywords" as the main metadata clues for the "bird and cat" and "town and temple" annotation, "data and time" for the "flower and food" annotation and "geographical coordinates" for the "lake and ocean" annotation.

---

[1] https://github.com/google/active-learning

Two volunteers in my Lab as annotators are involved in the image annotation experiment. Since one will be familiar with the images labeled by himself before, each of the annotators should label an image either with the plain interface or with our proposed CAIAF. In our experiment, each annotator labels two pairs with plain interface, and the other two pairs with CAIAF. We count their time used for labeling each batch of images as the annotation cost, and Figure 6 shows the results.

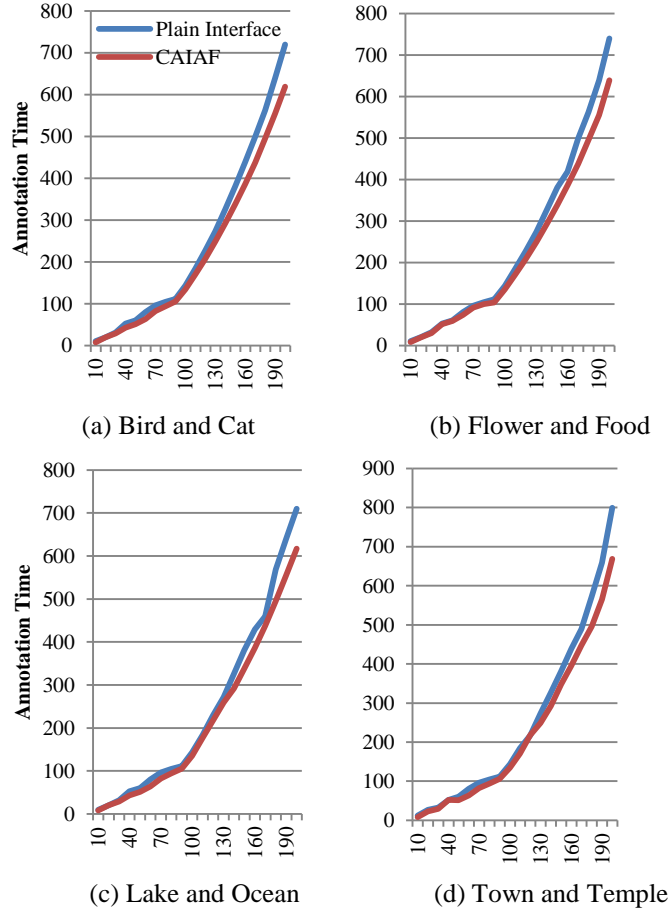

(a) Bird and Cat  (b) Flower and Food  (c) Lake and Ocean  (d) Town and Temple

**Fig. 6.** The comparison of CAIAF and conventional image annotation framework with plain interface without metadata clues and context in terms of cumulative annotation time.

From the result we can see that for all the four pairs of images, it takes less time for the annotation with CAIAF than plain framework as the annotation goes on. Images with useful context information yield to better annotation performance than plain interface without any context clues. The annotation of "Town and Temple" takes more time on average than the other three pairs, but it still uses less time with CAIAF than one with conventional active learning framework. Also, images labeled by CAIAF have less or at least equal errors than the plain one. Table 1 shows the final classification result. We can see that the F1 score of classification with CAIAF is equal to the plain one for the first pair, but a little bit higher for the other three pairs.

**Table 1.** F1 score of classification result (%)

|  | Learning with plain interface | Learning with CAIAF | Improvement |
|---|---|---|---|
| Bird and Cat | 61.6% | 61.6% | 0% |
| Flower and Food | 63.7% | 63.9% | 0.3% |
| Lake and Ocean | 61.4% | 62.3% | 1.4% |
| Town and Temple | 58.3% | 59.7% | 0% |

## 5    Conclusions and Future Work

Given a large pool of unlabeled images, active learning provides a way to iteratively select the most informative unlabeled images to label. In practice, batch mode active learning, where a set of items is picked all at once to be labeled and then used to retrain the classifier, is most feasible because it does not require the model to be retrained after each individual selection and makes most efficient use of human labor for annotation.

In this paper, we explored the possibility of reducing the annotation cost while maintaining the active learning performance. The experiment shows that our proposed context aware image annotation framework with plentiful metadata clues takes less time for the oracle to label images than the plain one without any metadata information as context. Also, the classification performance of active learning with CAIAF is equal or better than the plain one. In the future, we will explore the combination of multiple dimensions of context information and make more efficient annotation method such as semi-automatic or automatic annotation framework. We are also interested in the possibility of reducing annotation cost in other domains such as text or emails.

### Acknowledgment

This work was supported by the Ohio Department of Higher Education, the Ohio Federal Research Network and the Wright State Applied Research Corporation under award WSARC-16-00530 (C4ISR: Human-Centered Big Data).

### References

1. De Figueirêdo HF, Lacerda Y, de Paiva A, Casanova M, de Souza BC.: PhotoGeo: a photo digital library with spatial-temporal support and self-annotation. Multimed Tools Appl 59, pp. 279–305 (2012)

2. Ionescu B, Radu A-L,Menéndez M, Müller H, Popescu A, Loni B.: Div400: a social image retrieval result diversification dataset. In: Proceedings of the 5th ACM Multimedia Systems Conference, pp. 29–34. ACM, New York, NY, USA (2014)
3. Viana, W., Bringel Filho, J., Gensel, J., Oliver, M.V. and Martin, H.,: PhotoMap–Automatic Spatiotemporal Annotation for Mobile Photos. In: International Symposium on Web and Wireless Geographical Information Systems, pp. 187-201. Springer, Berlin, Heidelberg (2007)
4. Firmino, A.A., de Souza Baptista, C., de Figueirêdo, H.F., Pereira, E.T. and Amorim, B.D.S.P.,: Automatic and semi-automatic annotation of people in photography using shared events. Multimedia Tools and Applications, 1-35 (2018)
5. De Andrade, D.O.S., Maia, L.F., de Figueirêdo, H.F., Viana, W., Trinta, F. and de Souza Baptista, C.,: Photo annotation: a survey. Multimedia Tools and Applications, 77(1), 423-457 (2018)
6. Druck, G., Settles, B. and McCallum, A.,: Active learning by labeling features. In: Proceedings of the 2009 Conference on Empirical Methods in Natural Language Processing: Volume 1-Volume 1, pp. 81-90. Association for Computational Linguistics (2009)
7. Ravi, S. and Larochelle, H.,: Meta-learning for batch mode active learning. In: Proceedings of 6th International Conference on Learning Representations (ICLR 2018) workshop (2018)
8. Settles, B., Craven, M. and Friedland, L.,: Active learning with real annotation costs. In: Proceedings of the NIPS workshop on cost-sensitive learning, pp. 1-10 (2008)
9. Amato, G., Falchi, F. and Gennaro, C.,: Fast image classification for monument recognition. Journal on Computing and Cultural Heritage (JOCCH), 8(4), p.18 (2015)
10. Li, Y., Crandall, D.J. and Huttenlocher, D.P.,: Landmark classification in large-scale image collections. In: 2009 IEEE 12th international conference on computer vision, pp. 1957-1964. IEEE (2009)
11. Johnson, J., Ballan, L. and Fei-Fei, L.,: Love thy neighbors: Image annotation by exploiting image metadata. In: Proceedings of the IEEE international conference on computer vision, pp. 4624-4632 (2015)
12. Guillaumin, M., Verbeek, J. and Schmid, C.,: Multimodal semi-supervised learning for image classification. In 2010 IEEE Computer society conference on computer vision and pattern recognition, pp. 902-909. IEEE (2010)
13. Hwang, S.J. and Grauman, K.,: Learning the relative importance of objects from tagged images for retrieval and cross-modal search. International journal of computer vision, 100(2), pp.134-153 ( 2012)
14. Hays, J. and Efros, A.A.,: IM2GPS: estimating geographic information from a single image. In 2008 IEEE conference on computer vision and pattern recognition, pp. 1-8. IEEE (2008)
15. Roshan Zamir, A., Ardeshir, S. and Shah, M.,: Gps-tag refinement using random walks with an adaptive damping factor. In: Proceedings of the IEEE Conference on Computer Vision and Pattern Recognition, pp. 4280-4287 (2014)
16. Kim, G., Xing, E.P. and Torralba, A.,: Modeling and analysis of dynamic behaviors of web image collections. In European Conference on Computer Vision, pp. 85-98. Springer, Berlin, Heidelberg (2010)
17. Lindstaedt, S., Pammer, V., Mörzinger, R., Kern, R., Mülner, H. and Wagner, C.,: Recommending tags for pictures based on text, visual content and user context. In 2008 Third International Conference on Internet and Web Applications and Services, pp. 506-511. IEEE (2008)

18. Sigurbjörnsson, B. and Van Zwol, R.,: Flickr tag recommendation based on collective knowledge. In: Proceedings of the 17th international conference on World Wide Web, pp. 327-336. ACM (2008)
19. Sawant, N., Datta, R., Li, J. and Wang, J.Z.,: Quest for relevant tags using local interaction networks and visual content. In: Proceedings of the international conference on Multimedia information retrieval, pp. 231-240. ACM (2010)
20. Stone, Z., Zickler, T. and Darrell, T.,: Autotagging facebook: Social network context improves photo annotation. In 2008 IEEE computer society conference on computer vision and pattern recognition workshops, pp. 1-8. IEEE (2008)
21. McAuley, J. and Leskovec, J.,: Image labeling on a network: using social-network metadata for image classification. In European conference on computer vision, pp. 828-841. Springer, Berlin, Heidelberg (2012)
22. Luo, J., Boutell, M. and Brown, C.,: Pictures are not taken in a vacuum-an overview of exploiting context for semantic scene content understanding. IEEE Signal Processing Magazine, 23(2), pp.101-114 (2006)
23. Kalogerakis, E., Vesselova, O., Hays, J., Efros, A.A. and Hertzmann, A.,: Image sequence geolocation with human travel priors. In 2009 IEEE 12th international conference on computer vision, pp. 253-260. IEEE (2009)
24. Tang, K., Paluri, M., Fei-Fei, L., Fergus, R. and Bourdev, L.,: Improving image classification with location context. In: Proceedings of the IEEE international conference on computer vision, pp. 1008-1016 (2015)
25. Boutell, M. and Luo, J.,: Photo classification by integrating image content and camera metadata. In: Proceedings of the 17th International Conference on Pattern Recognition, ICPR 2004. Vol. 4, pp. 901-904. IEEE (2004)
26. Settles, B.,: Closing the loop: Fast, interactive semi-supervised annotation with queries on features and instances. In: Proceedings of the conference on empirical methods in natural language processing, pp. 1467-1478. Association for Computational Linguistics (2011)
27. Wei, Q., Franklin, A., Cohen, T. and Xu, H.,: Clinical text annotation–what factors are associated with the cost of time?. In AMIA Annual Symposium Proceedings, Vol. 2018, p. 1552. American Medical Informatics Association (2018)
28. Räbiger, S., Saygın, Y. and Spiliopoulou, M.,: How does tweet difficulty affect labeling performance of annotators?. arXiv preprint arXiv:1808.00388 (2018)
29. Baldridge, J. and Osborne, M.,: Active learning and the total cost of annotation. In Proceedings of the 2004 Conference on Empirical Methods in Natural Language Processing pp. 9-16 (2004)
30. Culotta, A. and McCallum, A.,: Reducing labeling effort for structured prediction tasks. In AAAI, Vol. 5, pp. 746-751 (2005)
31. Haertel, R., Felt, P., Ringger, E. and Seppi, K.,: Parallel active learning: eliminating wait time with minimal staleness. In: Proceedings of the NAACL HLT 2010 Workshop on Active Learning for Natural Language Processing, pp. 33-41. Association for Computational Linguistics (2010)
32. Cardoso, T.N., Silva, R.M., Canuto, S., Moro, M.M. and Gonçalves, M.A.,: Ranked batch-mode active learning. Information Sciences, 379, pp.313-337 (2017)
33. Sun, Y., Li, Q.: The research situation and prospect analysis of meta-search engines. In: 2nd International Conference on Uncertainty Reasoning and Knowledge Engineering, pp. 224-229. IEEE, Jalarta, Indonesia (2012).
34. Sun, Y., Loparo, K.: Context Aware Image Annotation in Active Learning with Batch Mode. In: 43rd Annual Computer Software and Applications Conference (COMPSAC), vol. 1, pp. 952-953. IEEE (2019).

26235. Li, Q., Zou, Y., Sun, Y.: Ontology based user personalization mechanism in meta search engine. In: 2nd International Conference on Uncertainty Reasoning and Knowledge Engineering, pp. 230-234. IEEE, Jalarta, Indonesia (2012).
36. Li, Q., Sun, Y., Xue, B.: Complex query recognition based on dynamic learning mechanism. Journal of Computational Information Systems 8(20), 8333-8340 (2012).
37. Li, Q., Zou, Y., Sun, Y.: User Personalization Mechanism in Agent-based Meta Search Engine. Journal of Computational Information Systems 8(20), 1-8 (2012).
38. Li, Q., Sun, Y.: An agent based intelligent meta search engine. In: International Conference on Web Information Systems and Mining, pp. 572-579. Springer, Berlin, Heidelberg (2012).
39. Sun, Y., Loparo, K.: Learning - based Adaptation Framework for Elastic Software Systems. In: Proceedings of 31st International Conference on Software Engineering & Knowledge Engineering, pp. 281-286. IEEE, Lisbon, Portugal (2019).
40. Sun, Y., Loparo, K.: A Clicked-URL Feature for Transactional Query Identification. In 2019 IEEE 43rd Annual Computer Software and Applications Conference (COMPSAC), vol. 1, pp. 950-951. IEEE (2019).
41. Sun, Y., Loparo, K.: Topic Shift Detection in Online Discussions using Structural Context. In 2019 IEEE 43rd Annual Computer Software and Applications Conference (COMPSAC) , vol. 1, pp. 948-949. IEEE (2019).
42. Sun, Y., Loparo, K.: Information Extraction from Free Text in Clinical Trials with Knowledge-Based Distant Supervision. In: IEEE 43rd Annual Computer Software and Applications Conference (COMPSAC), vol. 1, pp. 954-955. IEEE (2019).
43. Chua, T.S, Tang, J., Hong, R., Li, H., Luo, Z. and Zheng, Y.,: NUS-WIDE: a real-world web image database from National University of Singapore. In Proceedings of the ACM international conference on image and video retrieval, p. 48. ACM (2009)
44. Gong, Y., Jia, Y., Leung, T., Toshev, A. and Ioffe, S.,: Deep convolutional ranking for multilabel image annotation. arXiv preprint arXiv:1312.4894 (2013)
View publication stats